\documentclass[preprint,12pt,authoryear]{elsarticle}

\usepackage{amssymb}
\usepackage{amsmath}
\usepackage{hyperref}
\usepackage{booktabs}
\usepackage{xcolor}

\journal{Nuclear Physics B}
\usepackage{lineno}
\begin{document}

\begin{frontmatter}

\title{Towards Robust Deep Learning-based Rumex Obtusifolius Detection from Drone Images} 

\author[label1]{Fabian Dionys Schrag}
\author[label2]{Mehmet Ozgur Turkoglu}
\author[label3]{Konrad Schindler}
\author[label1]{Ralph Lukas Stoop}

\affiliation[label1]{organization={Agroscope NBA},
            addressline={Tänikon 1}, 
            city={Ettenhausen},
            postcode={8356}, 
            state={TG},
            country={Switzerland}}
\affiliation[label2]{organization={Agroscope  Earth Observation of Agroecosystems Team},
            addressline={Reckenholz 191}, 
            city={Zurich},
            postcode={8046}, 
            state={ZH},
            country={Switzerland}}
\affiliation[label3]{organization={ETH Zurich},
            addressline={Stefano-Franscini-Platz 5}, 
            city={Zurich},
            postcode={8093}, 
            state={ZH},
            country={Switzerland}}

\begin{abstract}
Domain adaptation (DA) addresses the challenge of transferring a machine learning model trained on a \emph{source domain} to a \emph{target domain} with a different data distribution. In this work, we study DA for the task of Rumex obtusifolius (Rumex) image classification. We train models on a published, ground vehicle-based dataset (source) and evaluate their performance on a custom target dataset acquired by unmanned aerial vehicles (UAVs). We find that Convolutional Neural Network (CNN) models, specifically ResNets, generalize poorly to the target domain, even after fine-tuning on the source data. Applying moment-matching and maximum classifier discrepancy, two established DA techniques, substantially improves target-domain performance. However, Vision Transformer (ViT) models pretrained with self-supervised objectives (DINOv2, DINOv3) handle domain shifts intrinsically well, surpassing even moment-matching-trained ResNets, likely due to the rich, general-purpose representations acquired during large-scale pretraining.  
Using ViTs fine-tuned on the source dataset, we demonstrate high classification performances in the range of $F_1=0.8$ on our target dataset. 
To support further research on DA for weed detection in grassland systems, we publicly release our UAV-based target dataset \textit{AGSMultiRumex}, comprising data from 15 flights over Swiss meadows.
\end{abstract}


\end{frontmatter}

\section{Introduction}
\label{sec1}
Rumex obtusifolius (Rumex), commonly known as broad-leaved dock, is a perennial weed with low feeding value. Native to Europe, it has become widespread over all temperate continents. In grassland systems, Rumex competes with desirable forage plants for resources like water or sunlight. Its high seed production may lead to a rapid infestation of a meadow, reducing usable grazing area, forage quality and yield \cite{OswaldHaggar1983}. To prevent this, current agricultural practice often involves the extensive use of herbicide, applied non-selectively to the whole meadow.\\
\\
Towards a more selective and resource-efficient control of Rumex in grassland systems, vision-based detection methods have gained a lot of interest. Their distinct visual features, including a large average size, made Rumex one of the best studied object of weed detection in grassland. In the mid-2000s, multiple studies have investigated classical computer vision for Rumex detection \cite{Gebhardt2006, Gebhardt2007, Evert2009}. However, these methods remained highly sensitive to image acquisition parameters and therefore could not be applied directly in practice. The rise of deep learning-based vision models has opened the door for automated detection systems for Rumex. Early work based on Convolutional Neural Networks (CNNs) to detect plants from camera images has demonstrated the supremacy of learning-based methods compared to conventional, rule-based vision algorithms \cite{Kounalakis2019}. Valente et al.~\cite{Valente2019, Valente2022} showed that the learning-based approach could also be employed for Rumex detection from images acquired by unmanned aerial vehicles (UAVs) and studied the detection performance for different flight altitudes. Lam et al.~\cite{Lam2020} proposed an end-to-end UAV workflow for weed detection, including image acquisition, orthomosaic generation, annotation, tiling, and CNN-based classification. Recently, RumexWeeds, a large dataset of high resolution images with Rumex plants was released \cite{Güldenring2023}. That data was captured at multiple sites in Denmark, using a standard camera mounted on a Clearpath Husky ground robot, moreover \cite{Güldenring2023} have reported detection scores with different YOLOX variants \cite{Ge2021}.\\
\\
Despite these advances, important challenges remain for practical, operational use of such computer vision models. Most published models were trained with the help of a dedicated dataset, specifically collected for the envisioned task. This is particularly true for the agricultural domain, where publicly available data that suits a specific target task is often scarce. For images, the standard workflow starts with a pretrained model (e.g., ResNet pretrained on ImageNet), which is then fine-tuned on a dataset specifically collected and annotated for the envisioned task. Similarly, published work on Rumex detection has focused on acquiring images from one or few meadows, at few (often only a single) points in time. Hence, the collected data is very specific and heavily biased towards specific image acquisition hardware, lighting conditions, types of meadows and the vegetational status. Consequently, they may not be suitable as a basis to develop models that operate under different conditions, even for the same application of Rumex detection or classification. Indeed, recent work \cite{Eichhorn2025} has shown that YOLO models fine-tuned on one of the three aforementioned datasets \cite{Valente2022, Güldenring2023, Eichhorn2025} fail to generalize to the other two, reaching $mAP50$ values of barely 30\%. This leads to the question how existing, already annotated images from one study can be effectively used to develop models for a specific agricultural context and application scenario.\\
\\
In machine learning theory, this setting is referred to as \textit{domain adaptation} (DA) \cite{BenDavid2010, farahani2020}. A domain $\mathcal{D} = (p, f)$ can be defined by a probability distribution $p$ over the input space and a labeling function $f$. DA investigates how a model trained on a source domain $\mathcal{D}_S$ can be transferred to a target domain $\mathcal{D}_T$. Under the common assumption of a domain-invariant labeling function, $f_T = f_S$, the objective is to address covariate shift arising from a change in the input distribution, that is, $p_T \neq p_S$.
With the rise of deep learning, DA has attracted substantial attention \cite{farahani2020}. Deep neural networks have high representational capacity and therefore require large amounts of labeled data for effective training, making retraining in new contexts costly and time-consuming. Moreover, their large number of parameters enables them to model even subtle characteristics of the training distribution, which increases the risk of overfitting to the source domain \cite{zhang_2017}. Consequently, DA is particularly relevant for deep networks, and numerous specialized methods have been proposed, see \cite{farahani2020} for a comprehensive review. To the best of our knowledge, however, these techniques have not yet been widely adopted in the agricultural domain.\\
\\
In this work, we investigate domain adaptation for Rumex image classification. We train several models on \textit{RumexWeeds} \cite{Güldenring2023}, a dataset of Rumex plants collected by a ground robot on Danish meadows, and evaluate their performance on a new, custom dataset, \textit{AGSMultiRumex}, which comprises images acquired during 15 individual UAV flights. This setup induces a pronounced domain shift due to differences in viewpoint, sensor characteristics, and acquisition conditions.
When using pretrained neural networks without any adaptation to the agricultural context, classification performance on the target dataset is consistently poor across all considered architectures. Fine-tuning on the RumexWeeds source dataset improves results, but ResNet-based predictors~\cite{he2015} still fail to generalize satisfactorily to the UAV domain. Their performance can, however, be substantially improved through appropriate (multi-source) domain adaptation, in particular with Moment Matching for Multi-Source Domain Adaptation ($\text{M}^3\text{SDA}-\beta$)~\cite{peng2019moment}.
In contrast, large vision transformer (ViT)~\cite{dosovitskiy2020image} models exhibit markedly stronger cross-domain generalization. When fine-tuned on the source dataset, they achieve high classification performance on AGSMultiRumex without requiring explicit domain adaptation. Notably, a DINOv3~\cite{simeoni_2025} fine-tuned on RumexWeeds using parameter-efficient low-rank adaptation (LoRA)~\cite{hu2021} attains a median $F_1$ score of $0.81$ across all UAV flights.
We attribute this improved transfer performance primarily to large-scale self-supervised pretraining rather than to architectural differences alone. The rich and diverse representations learned during pretraining appear to provide substantial robustness to domain shift, enabling effective reuse of existing labeled data. Our results, therefore, suggest that leveraging large, pretrained ViT models may currently be the most effective strategy to enhance generalization in agricultural vision tasks, potentially reducing or even eliminating the need for dedicated domain adaptation techniques.

\section{Materials and Method}
\subsection{RumexWeeds Dataset}
\begin{figure}[]
\centering
\includegraphics[width=\textwidth]{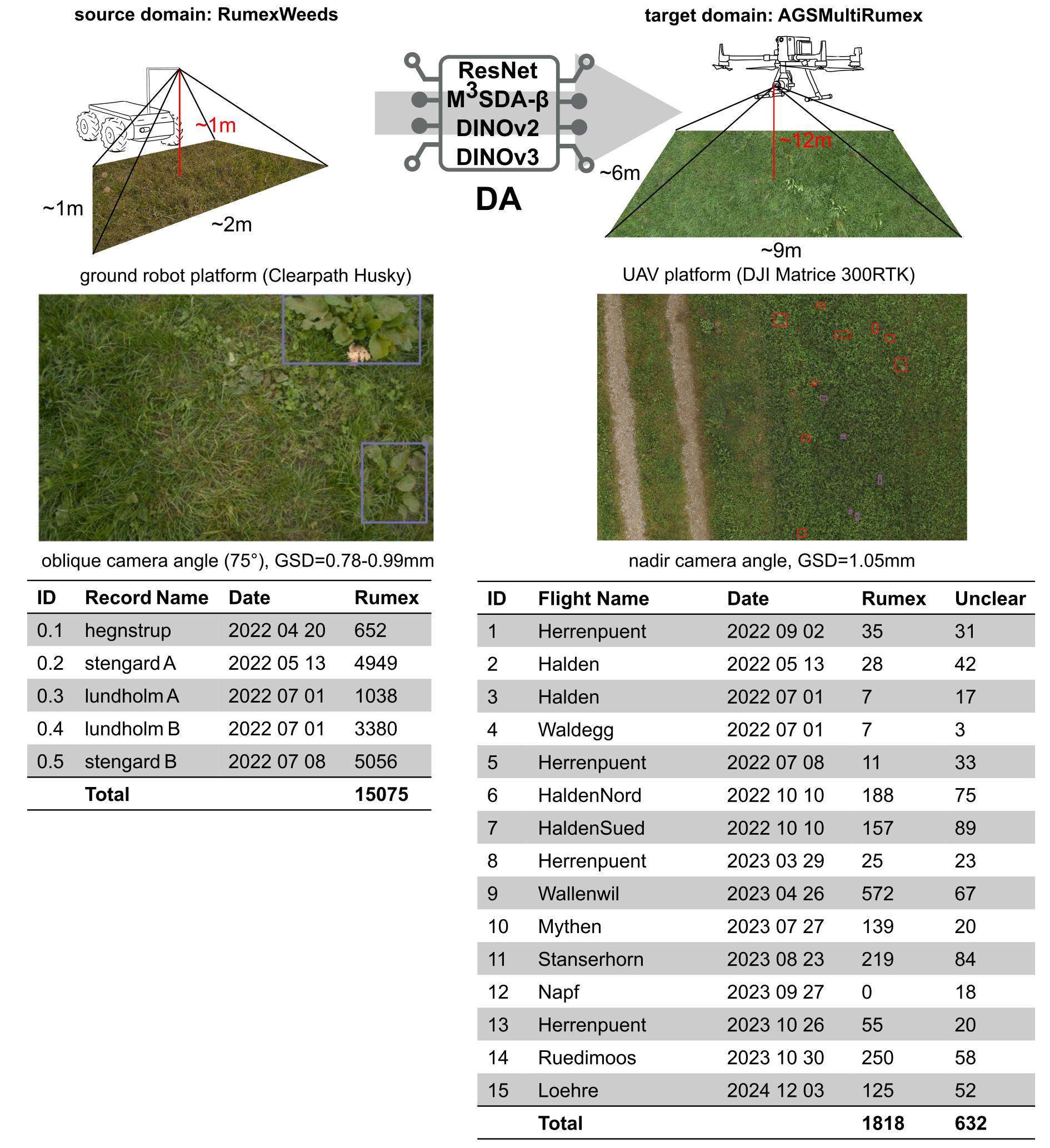}
\caption{Experimental setup with ground-robot-based source domain (left) used for training and UAV-based target domain (right) used for model validation. Different approaches are used to learn models based on the source domain to achieve robust detection in the target domain.}\label{fig1}
\end{figure}

For model training, we use RumexWeeds, an object detection dataset collected on three different dairy farms in Copenhagen, on four separate acquisition days during the summer and autumn of 2021, see Fig.~\ref{fig1} (left). Image acquisition is based on a Basler ace 2 RGB camera (Basler AG, Ahrensburg, Germany) mounted on a Clearpath Husky ground robot (Clearpath Robotics, Inc., Ontario, Canada)). Due to the camera's position at $1\,$m height above ground and an oblique angle of $75^\circ$, the imaging region covers a trapezoidal area of the ground. The resulting images have a resolution of 1920 × 1200 px and a ground sampling distance around 1mm. This setup results in high-resolution imagery of individual plants, where leaves and stems are clearly visible and rarely occluded.

\subsection{AGSMultiRumex Dataset}
To systematically evaluate domain adaptation methods under realistic conditions, we introduce \textit{AGSMultiRumex}, a novel validation dataset designed for robust Rumex detection. The dataset comprises 15 UAV flights over Swiss meadows conducted between April 2022 and December 2024, see Fig.~\ref{fig1} (right). 
All images were captured with a Zenmuse P1 camera (DJI, Shenzhen, China) mounted on a Matrice 300 RTK UAV (DJI, Shenzhen, China) in nadir configuration. The camera provides a native resolution of $8192 \times 5460$ pixels. Flights were performed at a fixed above-ground level of $AGL = 12\,\mathrm{m}$ using a $50\,\mathrm{mm}$ focal length lens, resulting in a ground sampling distance of $0.105\,\mathrm{cm/px}$ and an image footprint of approximately $9 \times 6\,\mathrm{m}^2$. The ground sampling distance is comparable to that of \textit{RumexWeeds}, making additional resolution harmonization unnecessary.
AGSMultiRumex covers meadows with highly diverse vegetational compositions, including dandelion, various herb species, and remnants of different crops or vegetables from previous rotations. In addition to this broad botanical variability, the dataset contains typical landscape elements frequently observed in UAV imagery of grasslands, such as streets, footpaths, fences, and vehicles. The inclusion of such structures enables a more realistic assessment of model robustness and out-of-distribution generalization. 
The dataset is publicly available at doi.org/10.5281/zenodo.18662099.

\subsection{Tiling-Based Reformulation of Rumex Detection as Image Classification}
\begin{figure}[]
\centering
\includegraphics[width=0.5\textwidth]{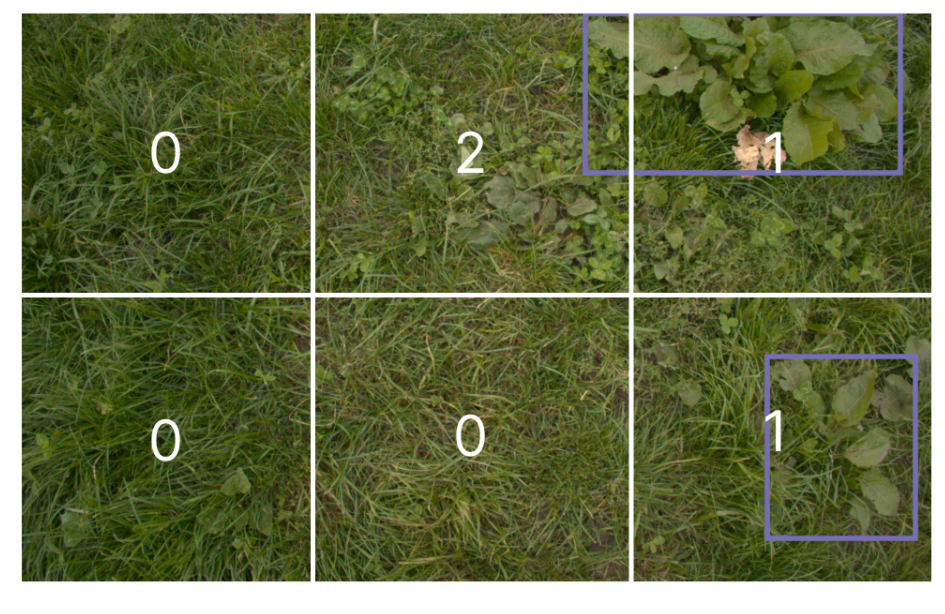}
\caption{Example demonstrating the translation from the original dataset used for object detection to the classification dataset. Label $0$ ($1$) depicts the negative (positive) samples containing no Rumex (containing Rumex). Label '2' indicates unclear labels due to area fraction of the positive label that are below a threshold ($r_{th}=0.1$).}\label{fig2}
\end{figure}

We cast Rumex detection as an image classification task in order to enable the direct application of established domain adaptation methods such as $\text{M}^3\text{SDA}$~\cite{peng2019moment}. To this end, each original image is partitioned into $N$ fixed-size tiles. We use a tile size of $518 \times 518$ pixels, corresponding to a footprint of approximately $0.5\,\text{m}^2$. This size provides sufficient spatial context to reliably capture Rumex plants while maintaining a favorable ratio between foreground and background pixels. It further represents a reasonable compromise between contextual information and computational efficiency.
To ensure full spatial coverage of images with arbitrary dimensions, we adopt a sliding-window strategy consisting of four separate passes, each initialized at one of the image corners (top-left, top-right, bottom-left, bottom-right). If the image dimensions are not exact multiples of the tile size, this procedure results in partially overlapping tiles. During training, such overlaps implicitly act as data augmentation, as the same plant may appear in slightly shifted spatial configurations across different tiles.
To derive tile-level labels from the original bounding box annotations, we define the normalized overlap ratio
\[
r = \frac{\text{bounding box pixel area} \cap \text{tile pixel area}}{\#\text{ tile pixels}}.
\]
Each tile is assigned the label `0' (background) if it does not overlap with any Rumex bounding box ($r = 0$), and `1' (Rumex) if the overlap exceeds a predefined threshold, $r > r_{th}$. Tiles with partial but insufficient overlap, $0 < r < r_{th}$, are labeled `2' (unclear) and excluded from model training. An example of the tiling procedure and the corresponding label assignment is shown in Fig.~\ref{fig2}.

\subsection{Dataset Splits}
To evaluate domain adaptation performance in an agricultural setting, we use \textit{RumexWeeds} as the source dataset for supervised training of models that classify image tiles as Rumex or background. The dataset comprises images collected at five distinct locations and dates. 
For single-source domain adaptation ($\text{M}^2\text{S}^2\text{DA}$) \cite{peng2019moment}, all five subsets are pooled into a single aggregated source domain. For multi-source domain adaptation ($\text{M}^3\text{SDA}-\beta$)~\cite{peng2019moment}, each location/date subset is treated as an individual source domain. In both settings, we employ a fixed train–validation split, where each split contains samples from all locations and dates to ensure representative coverage.
Since the original RumexWeeds dataset consists of video sequences recorded across multiple meadows and days, special care is taken to prevent data leakage. The splits are constructed such that no individual plant appears in more than one split, thereby avoiding spatial or temporal overlap between training and validation data.\\

\begin{figure}[]%% placement specifier
\centering%% For centre alignment of image.
\includegraphics[width=\textwidth]{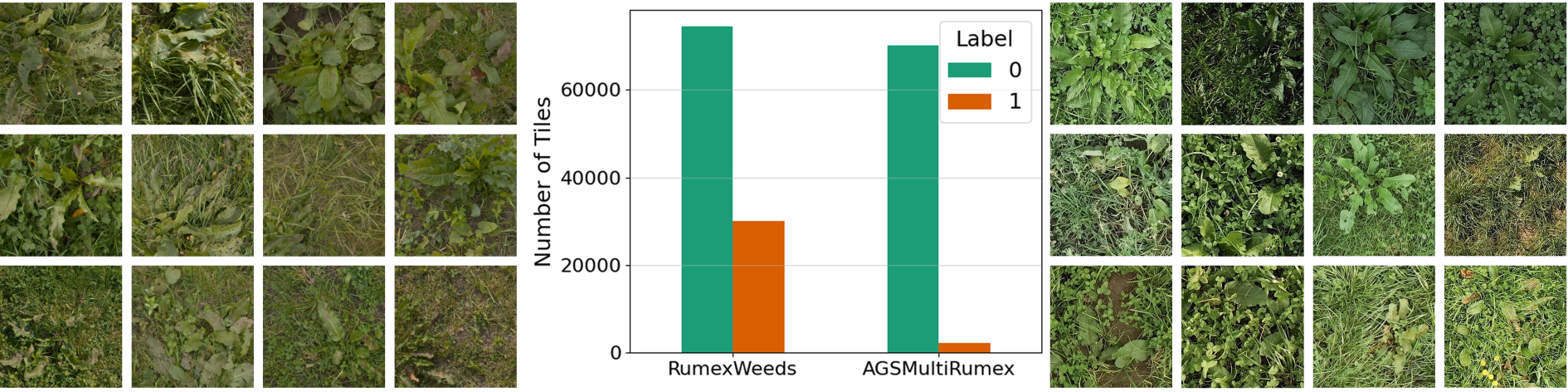}
\caption{Label distribution of the Rumex classification task in source and target data.}\label{fig3}
\end{figure}

To assess model performance under domain shift, we use \textit{AGSMultiRumex} as the target domain. The dataset comprises imagery from 15 UAV flights that vary in geographic location, vegetation composition, weather, and illumination conditions, thereby introducing substantial distributional differences relative to the source data. Although each flight could in principle be modeled as a separate domain or subdomain, we treat all flights jointly as a single target domain for simplicity.
A comparison of the resulting tile-level classification datasets is provided in Fig.~\ref{fig3}. While the number of negative (background) tiles is comparable between \textit{RumexWeeds} and \textit{AGSMultiRumex}, the source dataset contains substantially more positive (Rumex) tiles than the target dataset, leading to a more pronounced class imbalance in the latter.

\subsubsection{Model Architectures}
All considered models follow a modular architecture consisting of a feature extractor (backbone) $G$ and a classification head $C$. The backbone $G$ maps an input tile $x$ to a high-dimensional feature representation $z = G(x)$, while the head $C$ projects $z$ to class logits.
The classification head $C$ is implemented as a lightweight multilayer perceptron (MLP). Concretely, it comprises two fully connected layers. The first linear layer preserves the feature dimensionality of the backbone output and is followed by a ReLU activation and dropout with rate $p = 0.3$ for regularization. The second linear layer maps the intermediate representation to the two-dimensional output space corresponding to the binary classes (Rumex vs.\ background). This simple design isolates the representational capacity of the backbone and enables a controlled comparison across different feature extractors.\\

\paragraph{ResNet Backbones}

As convolutional baselines, we employ Residual Networks (ResNets) \cite{he2015}. ResNets are deep convolutional neural networks that employ residual connections to mitigate vanishing gradient effects and facilitate the optimization of very deep architectures. Due to their strong inductive bias for locality and translation equivariance, ResNets remain widely used as generic visual feature extractors and serve as standard backbones in many detection frameworks, such as Fast(er) R-CNN~\cite{girshick2015fast, ren_faster_2016}.
All ResNet models are initialized with weights pretrained on ImageNet~\cite{deng2009imagenet}, providing generic visual features prior to adaptation to the agricultural domain. We investigate multiple capacities, namely ResNet50, ResNet101, and ResNet152, to analyze the influence of network depth and representational capacity on cross-domain generalization performance.

\paragraph{DINOv2 and DINOv3}
To complement convolutional baselines, we evaluate transformer-based~\cite{vaswani2017attention} feature extractors pretrained with large-scale self-supervised learning, namely DINOv2 \cite{oquab2024} and DINOv3 \cite{simeoni_2025}. Both approaches build upon the Vision Transformer (ViT) architecture \cite{dosovitskiy2021} and are trained using a self-distillation objective without manual annotations, enabling the learning of semantically rich and transferable visual representations from massive collections of unlabeled images.
In contrast to supervised pretraining, the self-supervised DINO framework encourages invariance against a larger class of appearance variations (respectively, augmentations) and promotes the emergence of high-level semantic structure in the learned embedding space. This property is particularly relevant under domain shift, where robustness and generalization beyond the training distribution are critical.
We consider DINO-based feature extractors at three different model scales, denoted as \textit{small}, \textit{base}, and \textit{large}, in order to analyze the effect of model capacity and representation quality on downstream classification and cross-domain transfer performance.

\subsection{Training and Adaptation Strategies}

\paragraph{Supervised Training (Vanilla)}
The vanilla training strategy corresponds to standard supervised learning on the labeled source dataset. The backbone $G$ and classifier $C$ are combined and optimized jointly using the cross-entropy loss.
For all models, we consider two principal regimes: (i) \textit{frozen}, where the backbone $G$ remains fixed and only the classifier $C$ is trained, and (ii) \textit{fine-tuning} (FT), where parts of $G$ are unfrozen and updated during training. The degree of fine-tuning is controlled by the parameter \textit{unfreeze}, which specifies the number of trainable backbone blocks.
For ResNet-based models, \textit{unfreeze} = 10 corresponds to full fine-tuning, independent of network depth. In the main text, we report results for \textit{unfreeze} $\in \{0, 10\}$, while intermediate configurations (\textit{unfreeze} $\in \{3, 6\}$) are provided in the appendix (Fig. \ref{figSI2}).
For DINO-based Vision Transformers, the total number of transformer blocks depends on model size, namely 12, 18, and 24 blocks for \textit{small}, \textit{base}, and \textit{large}, respectively. We evaluate DINO \textit{small} and \textit{base} for \textit{unfreeze} $\in \{0, 6, 12\}$ and DINO \textit{large} for \textit{unfreeze} $\in \{0, 6, 12, 18\}$. Larger configurations are constrained by available computational resources.\\

\paragraph{Unsupervised Domain Adaptation}
For ResNet-based models, we additionally investigate unsupervised domain adaptation via moment matching. These approaches aim to align statistical properties of the source and target feature distributions, specifically their first and second moments, under the assumption that reducing distributional discrepancies in feature space facilitates transfer when the labeling function remains approximately invariant.
We consider both single-source ($\text{M}^2\text{S}^2\text{DA}$) and multi-source ($\text{M}^3\text{SDA}-\beta$) variants~\cite{peng2019moment}. In both cases, the standard supervised classification loss on labeled source samples is augmented with a regularization term that penalizes discrepancies between the mean and covariance of source and target feature representations. In the multi-source setting, the objective further includes alignment terms between the different source domains and a classifier discrepancy mechanism \cite{saito2018}. A detailed methodological description is provided in the \ref{app_DA}. \\

\paragraph{Low Rank Adaptation (LoRA)}
Fine-tuning large Vision Transformers such as DINOv2 and DINOv3 is memory- and compute-intensive. To enable efficient adaptation, we employ Low-Rank Adaptation (LoRA) \cite{hu2021}, a parameter-efficient fine-tuning (PEFT) technique \cite{xin2025}. 
LoRA freezes the original pretrained weights and introduces trainable low-rank decomposition matrices into selected linear layers. This approach substantially reduces the number of trainable parameters while retaining the representational capacity of the pretrained backbone. We evaluate ranks $R \in \{8, 16, 32\}$ for DINO models of all sizes to study the trade-off between adaptation capacity and computational efficiency.

\subsection{Evaluation Protocol}
Model performance is evaluated using the $F_1$ score,
\begin{equation}
F_1 = \frac{2\cdot TP}{2\cdot TP + FP + FN}\;\;,
\end{equation}
where $TP$ denotes the number of correctly classified Rumex tiles (true positives), $FP$ the number of background tiles incorrectly predicted as Rumex (false positives), and $FN$ the number of Rumex tiles missed by the model (false negatives).
During training, the $F_1$ score is computed on the target dataset separately for each UAV flight at every epoch. For flights containing only few Rumex instances, these per-flight scores can exhibit substantial epoch-to-epoch variability, as small changes in predictions strongly affect the metric (see Fig.~\ref{figSI1} in the appendix). Flight Napf (ID 12) is omitted for our analysis, since it does not contain any Rumex.\\

Model selection is performed after an initial warm-up period of five epochs. We select the epoch that maximizes the $F_1$ score on the RumexWeeds validation set. At this selected epoch, we report the mean $\bar{F}_1$ and median $\tilde{F}_1$ across all target flights. To quantify training-related variability, we additionally compute the standard deviation $\sigma_{\text{epochs}}$ of $\tilde{F}_1$ across the considered epochs.

\subsection{Implementation Details and Computational Resources}%
All experiments were conducted on Agroscope’s internal compute cluster, equipped with five NVIDIA A6000 GPUs, each providing 48\,GB of VRAM. Model training was implemented in PyTorch \cite{pytorch}, and experiment management, logging, and reproducibility were handled using \textit{MLflow}.
The most computationally demanding configuration, LoRA with rank 32 applied to DINOv3 \textit{large}, required approximately 72 hours for a full 90-epoch training run on a single GPU.

\section{Experimental Results}

\begin{figure}[]%% placement specifier
\centering%% For centre alignment of image.
\includegraphics[width=\textwidth]{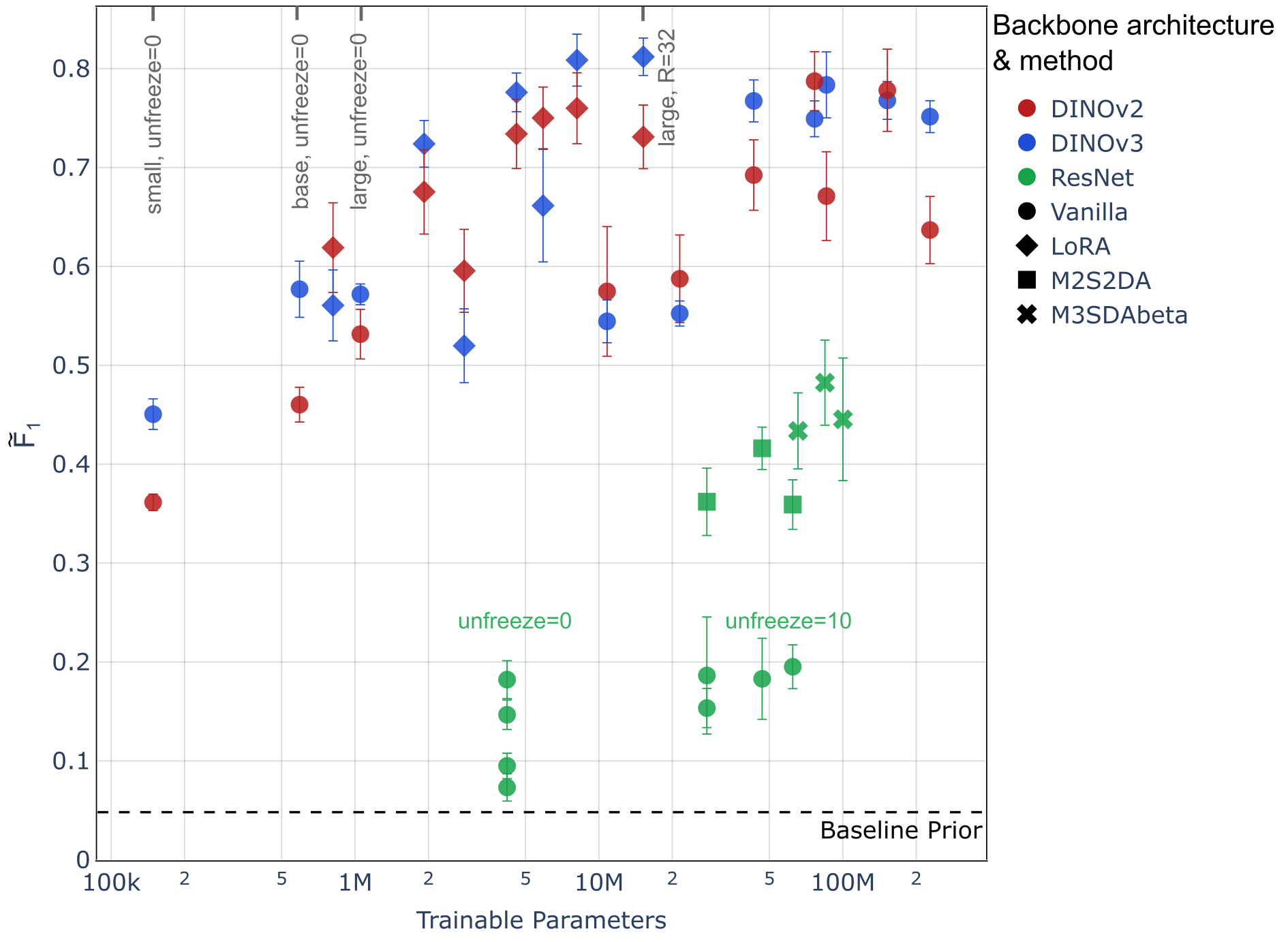}
\caption{Model performance on AGSMultiRumex vs. number of trainable parameters for different models using DINOv2 (red), DINOv3 (blue) or ResNet (green). ResNet-based models are trained either in a standard, vanilla way (circles) or using moment-matching DA in a single-source ($\text{M}^2\text{S}^2\text{DA}$, squares) or multi-source ($\text{M}^3\text{SDA}-\beta$, crosses) setting. DINO-based models are trained in a straightforward fashion (vanilla, circles) or using parameter-efficient low-rank approximations (LoRA, diamonds). Error bars correspond to the standard deviation of the median measured over the last 10 epochs, $\sigma_{epoch}$. The dashed line indicates the baseline of a dummy classifier based on prior Rumex probability.}\label{fig4}
\end{figure}

Our main results are summarized in Fig.~\ref{fig4}, which compares model performance across architectures and training strategies as a function of the number of trainable parameters. Performance is measured by the median $\tilde{F}_1$ score of the minority class (\textit{Rumex}) over all 15 AGSMultiRumex flights.

\paragraph{ResNet-based models}
The weakest target-domain performance is observed for ResNet backbones. When the feature extractor remains frozen (\textit{unfreeze}=0), ResNet models fail to reliably detect Rumex, largely independent of model depth or pretraining dataset (ImageNet or iNaturalist). These configurations (green circles around $\approx 4.2$M trainable parameters) achieve only $\tilde{F}_1\approx0.1$--$0.2$, albeit still above the dummy baseline based on class priors (dashed line).
Notably, even in the frozen setting, the number of trainable parameters is relatively high due to the 2048-dimensional feature embeddings produced by ResNet, which are mapped to the two output logits by the classifier head. \\

Full fine-tuning of the ResNet backbone (\textit{unfreeze} = 10) substantially improves performance on the RumexWeeds validation set (see Fig.~\ref{figSI2} in the appendix), but does not translate into better target-domain performance. The cluster of green points at $27$--$62$M parameters remains at $\tilde{F}_1 \approx 0.15$--$0.2$. This behavior is consistent across backbone sizes (ResNet50, 101, 152) and different degrees of unfreezing. While deeper fine-tuning improves in-domain validation performance (see Fig.~\ref{figSI2} in the appendix), it does not mitigate domain shift.\\

Substantial gains for ResNet-based models are only achieved when incorporating moment-matching domain adaptation. Single-source $\text{M}^2\text{S}^2\text{DA}$ increases performance to $\tilde{F}_1 \approx 0.36$, while multi-source $\text{M}^3\text{SDA}-\beta$ further improves results to $\tilde{F}_1 \approx 0.45$. These results demonstrate that explicit distribution alignment is necessary for convolutional backbones to generalize across agricultural domains. 
However, in the multi-source case the model effectively becomes an ensemble, and total parameter count scales with the number of source domains, reaching up to $\sim 100$M parameters for ResNet152 with $\text{M}^3\text{SDA}-\beta$. Within this framework, increasing backbone depth has only a minor influence on target performance.

\paragraph{DINO-based models}

The strongest performance is obtained with Vision Transformers. Even without fine-tuning ($unfreeze=0$), the smallest DINO models already achieve performance comparable to moment-matching ResNet approaches, despite having more than two orders of magnitude fewer trainable parameters. DINOv3 consistently outperforms DINOv2 in the frozen regime across model scales.
Increasing model size from \textit{small} to \textit{base} yields substantial gains, while performance saturates around $\tilde{F}_1 \approx 0.5$--$0.6$ for frozen large models. Further improvements are obtained through backbone adaptation. Both full fine-tuning and LoRA lead to marked gains, with the best results achieved by DINOv3 \textit{large} with LoRA ($R=32$), reaching $\tilde{F}_1 \approx 0.81$. No consistent trend is observed between DINOv2 and DINOv3 after fine-tuning, although the overall best configuration in our study is DINOv3 with LoRA .

\paragraph{Precision--Recall Analysis}
Tab.~\ref{table1} reports per-flight precision and recall for three representative models: frozen DINOv3-\textit{small}, ResNet101 with $\text{M}^3\text{SDA}-\beta$, and DINOv3-\textit{large} with LoRA.
Although the models based on DINOv3-\textit{small} and ResNet101 achieve similar median $\tilde{F}_1$, their error profiles differ substantially. The ResNet-based DA model attains relatively high recall but suffers from low precision, indicating frequent false positives. In contrast, the frozen DINO model achieves comparatively high precision but low recall, missing a substantial fraction of Rumex tiles.
\\
Fine-tuning significantly improves recall for ViT-based models while maintaining strong precision. The best-performing DINOv3-LoRA model achieves median and mean recall above 0.8, with precision also improving overall. Nevertheless, certain flights remain prone to false positives, limiting precision in challenging scenarios.

\begin{figure}[]%% placement specifier
\centering%% For centre alignment of image.
\includegraphics[width=\textwidth]{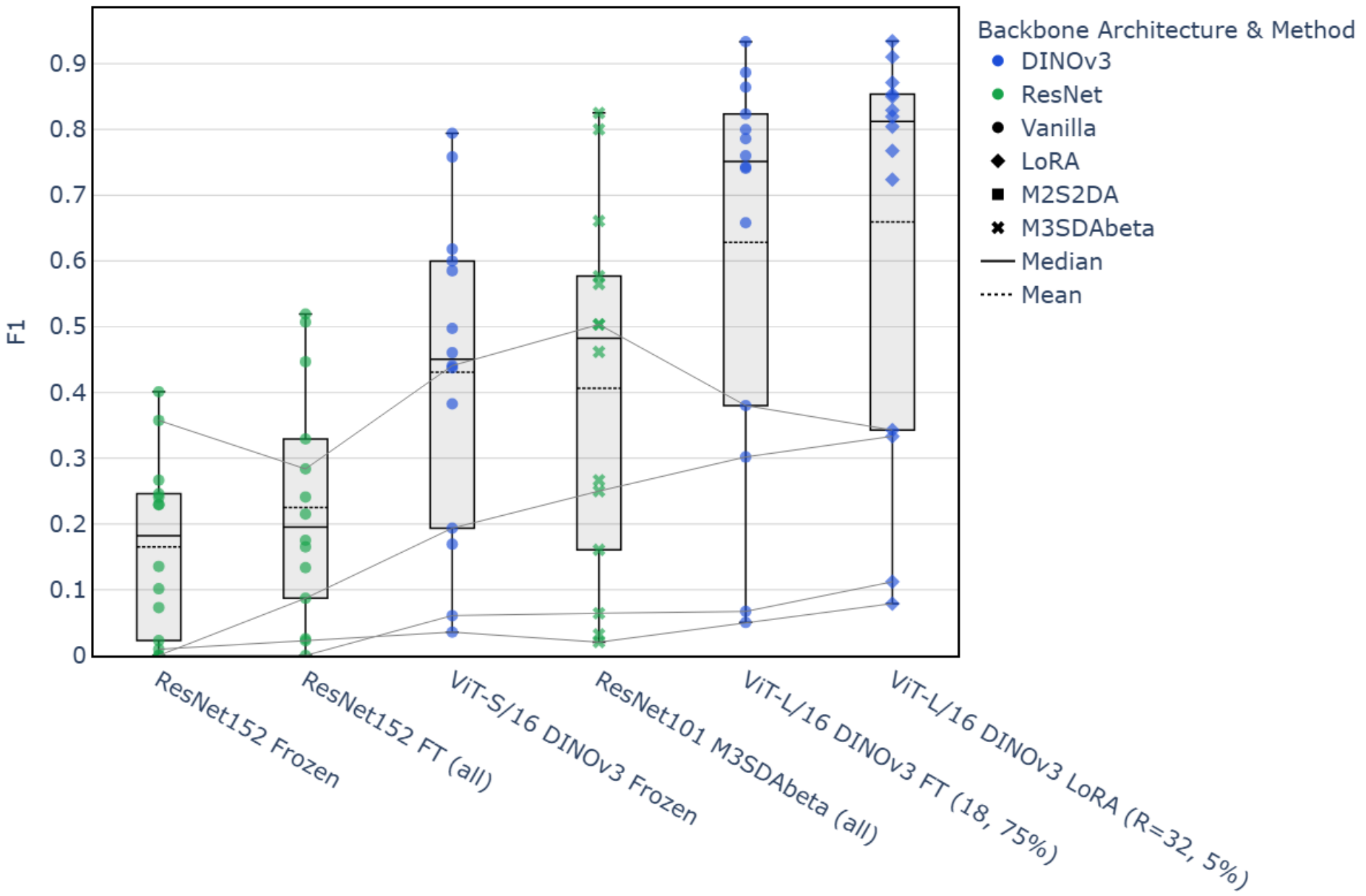}
\caption{$F_1$ score for each flight for compared methods.}\label{fig5}
\end{figure}

\paragraph{Per-Flight Performance}
Fig.~\ref{fig5} shows per-flight $F_1$ scores for selected models. Transitioning from convolutional to transformer-based backbones improves performance on the majority of flights. However, specific flights, including Halden (ID 3), Waldegg (ID 4), Herrenpuent (ID 5), and Herrenpuent (ID 13), remain challenging across all architectures and training strategies, indicating persistent domain-specific difficulties (see Tab.~\ref{table1}). Sample images of these flights together with model predictions are given in the appendix (Fig.~\ref{figSI4} and \ref{figSI3}).

\begin{table}
    \centering
    \begin{tabular}{l|cc|cc|cc}
    \toprule
        \textbf{Flight} 
        & \multicolumn{2}{c|}{\textbf{DINOv3-s}}
        & \multicolumn{2}{c|}{\textbf{ResNet101}}
        & \multicolumn{2}{c}{\textbf{DINOv3-l}} \\
                & \multicolumn{2}{c|}{\textbf{frozen}}
        & \multicolumn{2}{c|}{\textbf{$\text{M}^3\text{SDA}-\beta$}}
        & \multicolumn{2}{c}{\textbf{LoRA}} \\
         & prec & rec & prec & rec & prec & rec\\
         \midrule
        1: Herrenpuent & 0.79 & 0.09 & 0.65 & 0.52 & 1.00 & 0.67\\
        2: Halden & 0.37 & 0.40 & 0.02 & 0.93 & 0.86 & 0.69 \\
        3: Halden & 0.14 & 0.33 & 0.16 & 0.56 & 0.21 & 0.89 \\
        4: Waldegg & 0.02 & 0.75 & 0.01 & 1.0 & 0.04 & 1.0 \\
        5: Herrenpuent & 0.04 & 0.19 & 0.04 & 0.38 & 0.06 & 0.44 \\
        6: HaldenNord & 0.90 & 0.31 & 0.84 & 0.36 & 0.85 & 0.86 \\
        7: HaldenSued & 0.93 & 0.29 & 0.09 & 0.85 & 0.99 & 0.71 \\
        8: Herrenpuent & 0.55 & 0.67 & 0.32 & 0.83 & 0.77 & 0.94 \\
        9: Wallenwil & 0.92 & 0.47 & 0.87 & 0.78 & 0.98 & 0.85 \\
        10: Mythen & 0.39 & 0.69 & 0.16 & 0.81 & 0.58 & 0.97 \\
        11: Stanserhorn & 0.90 & 0.43 & 0.76 & 0.58 & 0.80 & 0.96 \\
        13: Herrenpuent & 0.30 & 0.83 & 0.37 & 0.78 & 0.21 & 1.0\\
        14: Ruedimoos & 0.99 & 0.66 & 0.98 & 0.40 & 0.97 & 0.90 \\
        15: Loehre & 0.98 & 0.62 & 0.94 & 0.70 & 0.89 & 0.76 \\
        \midrule
        \textit{Median} & \textit{0.67} & \textit{0.45} & \textit{0.35} & \textit{0.74} & \textit{0.82} & \textit{0.87} \\
        \textit{Mean} & \textit{0.58} & \textit{0.48} & \textit{0.44} & \textit{0.68} & \textit{0.66} & \textit{0.83} \\  
        $\sigma$ & \textit{0.36} & \textit{0.22} & \textit{0.37} & \textit{0.20} & \textit{0.35} & \textit{0.15} \\ 
        \bottomrule
    \end{tabular}
    \caption{Per-flight precision (prec) and recall (rec) comparison.}
    \label{table1}
\end{table}

\section{Discussion}

Our results demonstrate that modern large-scale vision models pretrained on publicly available data can transfer effectively to new agricultural settings, even under rather large domain shifts in terms of viewpoint, sensor characteristics, vegetation composition, and illumination. Despite pronounced differences between \textit{RumexWeeds} and \textit{AGSMultiRumex}, we achieve target-domain performance up to $\tilde{F}_1 = 0.81$. This highlights the practical value of reusing published agricultural datasets and leveraging strong pretrained representations.
In contrast, vanilla fine-tuned ResNet models fail to generalize to UAV imagery, consistent with recent findings in agricultural vision \cite{Eichhorn2025}. Explicit domain adaptation via $\text{M}^3\text{SDA}-\beta$ substantially improves performance for ResNet-based models, yet absolute $F_1$ scores remain moderate. A decisive improvement is obtained when replacing convolutional backbones with Vision Transformers. Off-the-shelf DINOv2 and DINOv3 feature extractors without any fine-tuning already match or surpass ResNet models trained with moment-matching DA. This strong zero-shot transfer likely stems from large-scale self-supervised pretraining on highly diverse image corpora, which yields semantically structured and robust feature spaces. Nevertheless, even these foundation models benefit markedly from adaptation to the agricultural domain. In particular, LoRA fine-tuning leads to substantial additional gains, in line with recent findings for DINO-pretrained models in agriculture~\cite{ESPEJOGARCIA2025110900}.\\

Except for the smallest ViT variants, transformer-based models consistently have an edge over ResNets, across different training strategies and repeated runs. Within the ViT family, performance differences are more nuanced. DINOv3 tends to slightly outperform DINOv2, although margins are generally small. Both full fine-tuning and LoRA lead to pronounced improvements, indicating that parameter-efficient fine-tuning is a viable alternative to updating all backbone weights. Performance further increases with model size, particularly when fine-tuned. Importantly, the LoRA approach allows scaling to large models with tractable computational cost.\\

A central challenge in our study is the substantial flight-to-flight variability within AGSMultiRumex. Per-flight analysis reveals that no single model performs uniformly well across all conditions. Even the best ViT-based models struggle on specific flights, notably Halden (ID 3), Waldegg (ID 4), Herrenpuent (ID 5), and Herrenpuent (ID 13). Qualitative inspection reveals two main failure modes.
First, lighting conditions and phenological stage significantly affect detection quality. For example, the flight Herrenpuent (ID 5) is characterized by very bright illumination and early vegetative growth stages, which reduce visual contrast between Rumex and surrounding vegetation, see Fig.~\ref{figSI4} in the appendix. This flight also exhibits a threefold higher number of ``unclear'' annotations compared to confirmed Rumex labels, indicating that it is inherently challenging even for human annotators. With the exception of this flight, however, recall remains consistently high for the best-performing ViT model, suggesting that the transferred representations are generally robust to UAV-specific imaging conditions.
Second, systematic false positives limit precision in certain flights. Large-leaf dandelions, which are underrepresented in \textit{RumexWeeds}, are frequently misclassified as Rumex in flights such as Waldegg (ID 4), Herrenpuent (ID 5), and Herrenpuent (ID 13), see Fig.~\ref{figSI3} in the appendix. This reflects a classical selection bias: if specific confounding species are rare in the source dataset, the model cannot learn a reliable decision boundary that separates them from the target class. Addressing this issue requires an extended training set with representative labeled examples. However, given the strong visual similarity between large dandelions and Rumex, even additional data may not completely resolve the ambiguity.\\

Surprisingly, false positives also arose from visually dissimilar objects, most notably a car in flight Halden (ID 3), which most models at least partially classified as \textit{Rumex}. One might attribute this to aggressive fine-tuning that degrades the backbone's original representations; however, we observe the same behavior even with frozen DINOv2/DINOv3 feature extractors. This is particularly unexpected, as DINO-pretrained backbones produce powerful features that should readily distinguish cars from meadow vegetation. The likely explanation is that the RumexWeeds training set contains no car instances, so the classifier never learns to separate cars from \textit{Rumex}, even when the underlying feature representations are clearly distinct.\\

Overall, our findings suggest that large, self-supervised ViT backbones provide a strong foundation for agricultural vision under domain shift. However, performance remains constrained by dataset bias, rare confounders, and extreme imaging conditions. Future improvements are therefore likely to require not only architectural advances, but also more diverse and carefully curated training data.

\section{Conclusions}
In this work, we demonstrate that SOTA ViT-based computer vision models can successfully be applied to DA tasks in the agricultural context. Based on our findings, this even holds for cases where the source and target images are acquired by very different platforms, such as a ground vehicle compared to UAV. While ResNet-based models need dedicated training routines facilitating DA, transformer-based models may intrinsically provide better generalization capabilities. For our studied task of Rumex detection, we demonstrate a median $\tilde{F_1}=0.81$ on our AGSMultiRumex validation set. These findings highlight the role of publicly available datasets that can lead to rapid prototype development, if model architecture is properly selected. 

\section*{Funding} This work was supported by the Innosuisse Swiss Innovation Agency [project 47541.1 IP-EE].

\section*{Credit authorship contribution statement} \textbf{Fabian Dionys Schrag:} Investigation, Data curation, Methodology, Formal analysis, Software, Writing - review and editing. \textbf{Mehmet Ozgur Turkoglu:} Conceptualization, Supervision, Writing - review and editing. \textbf{Konrad Schindler:} Conceptualization, Supervision, Writing - review and editing. \textbf{Ralph Lukas Stoop:} Investigation, Data curation, Conceptualization, Supervision, Writing - original draft.

\section*{Declaration of generative AI and AI-assisted technologies in the manuscript preparation process}

During the preparation of this work the author(s) used DeepL in order to enhance phrasing and overall language quality of the manuscript. After using this tool/service, the author(s) reviewed and edited the content as needed and take(s) full responsibility for the content of the published article.

\section*{Acknowledgment}
The authors are grateful to Markus Keller and Markus Sax from Agroscope for conducting the UAV flight campaigns and providing the raw aerial photographs.

\section*{Data availability}
The dataset is publicly available at
doi.org/10.5281/zenodo.18662099.

\bibliographystyle{elsarticle-harv}
\bibliography{references}

\clearpage

\appendix
\section{Domain Adaptation}\label{app_DA}
\paragraph{$\text{M}^2\text{S}^2\text{DA}$}
$\text{M}^2\text{S}^2\text{DA}$ (Moment Matching for Single Source Domain Adaptation) is a simplified version of $\text{M}^3\text{SDA}$ \cite{peng2019moment} assuming one labeled source domain and one unlabeled target domain. The method aligns the feature distributions of source and target domains by minimizing the differences between their first and second moments (mean and covariance) in the feature space:
$$
MD^2(\mathcal{D}_S, \mathcal{D}_T) = \sum_{k=1}^2\|\mathbb{E}(G(\mathbf{X}_S)^k) - \mathbb{E}(G(\mathbf{X}_T)^k)\|_2.
$$
Here $G$ is the feature extractor transforming input images $\mathbf{X}$ into to the feature space. The training objective combines the supervised binary cross entropy loss $\mathcal{L}_{CE}$ on the labeled source domain with the moment matching loss $MD^2(\mathcal{D}_S, \mathcal{D}_T)$ weighted by a factor $\lambda$:
$$
\min_{G, C} \mathcal{L}_{CE} + \lambda \min_G MD^2(\mathcal{D}_S, \mathcal{D}_T)
$$
where the total loss is minimized with respect to $G$ and the classifier head $C$. 
This training objective enforces the model to learn domain-invariant representations while still optimizing for classification on the labeled source domain. It represents a basic DA approach for single source-single target adaptation.

\paragraph{$\text{M}^3\text{SDA}-\beta$}
$\text{M}^3\text{SDA}$ \cite{peng2019moment} extends the moment matching idea of $\text{M}^2\text{S}^2\text{DA}$ to the multi-source setting, where labeled data from $N$ source domains are available. The method simultaneously aligns the moments across multiple sources and the target. The Moment Distance $MD^2(\mathcal{D}_S, \mathcal{D}_T)$ is extended to align each source with the target domain and all source domains with one another:
\begin{align}
MD^2(\mathcal{D}_S, \mathcal{D}_T) =& \sum_{k=1}^2\bigg(\frac{1}{N}\sum_{i=1}^N\|\mathbb{E}(G(\mathbf{X}_i)^k) - \mathbb{E}(G(\mathbf{X}_T)^k)\|_2  \\
&+{N \choose 2}^{-1}\sum_{i=1}^{N-1}\sum_{j=i+1}^N\|\mathbb{E}(G(\mathbf{X}_i)^k) - \mathbb{E}(G(\mathbf{X}_j)^k)\|_2\bigg)
\end{align}
The resulting training objective is given as:
$$
\min_{G, C} \sum_{i=1}^N \mathcal{L}_{D_i} + \lambda \min_G MD^2(\mathcal{D}_S, \mathcal{D}_T)
$$
where $\mathcal{L}_{D_i}$ refers to the cross entropy loss $\mathcal{L}_{CE}$ of the corresponding domain $D_i$.\\
\\
Furthermore, M2S2DA and $\text{M}^3\text{SDA}$ assume that conditional distributions $p(x|y)$ will align automatically once $p(x)$ is aligned, which is not always true. To address this, $\text{M}^3\text{SDA}-\beta$ additionally introduces a classifier discrepancy mechanism \cite{saito2018}: for each source domain, two classifiers ($C_i, C^\prime_i$) are trained. They are optimized to maximize their disagreement on target samples, while the feature extractor $G$ is optimized to minimize this discrepancy.
The training procedure of $\text{M}^3\text{SDA}-\beta$ consists of three steps. First, $G$ and $C^\prime = \{(C_1, C^\prime_1), ... , (C_N, C^\prime_N) \}$ are trained to classify the multi-source samples correctly:
$$
\min_{G, C^\prime} \sum_{i=1}^N \mathcal{L}_{D_i} + \lambda \min_G MD^2(\mathcal{D}_S, \mathcal{D}_T)
$$
Then, the feature extractor $G$ is fixed and each pair of the classifiers are trained such as the discrepancy of each pair is as large as possible on the target domain:
$$
\min_{C^\prime} \sum_{i=1}^N \mathcal{L}_{D_i} - \sum_i^N |P_{C_i}(D_T) - P_{C^\prime_i}(D_T)|
$$
Here, $P_{C_i}(D_T)$ and $P_{C_i\prime}(D_T)$ represent the outputs of $C_i$ and $C^\prime_i$ on the target domain.
The last step is to fix $C^\prime$ and train the feature extractor $G$ to minimize the discrepancy of each classifier pair on the target domain:
$$
\min_G \sum_i^N |P_{C_i}(D_T) - P_{C^\prime_i}(D_T)|
$$
By alternating these steps, $\text{M}^3\text{SDA}-\beta$ aligns both the covariate distributions $p(x)$ (via moment matching) and the conditional distributions $p(y|x)$ (via classifier discrepancy).
The above formulation and training strategy are adapted from the original $\text{M}^3\text{SDA}$ and $\text{M}^3\text{SDA}-\beta$ algorithms introduced by Peng et al. \cite{peng2019moment}.

\section{Validation $F_1$ scores during model training}
\begin{figure}[thb]%% placement specifier
\centering%% For centre alignment of image.
\includegraphics[width=\textwidth]{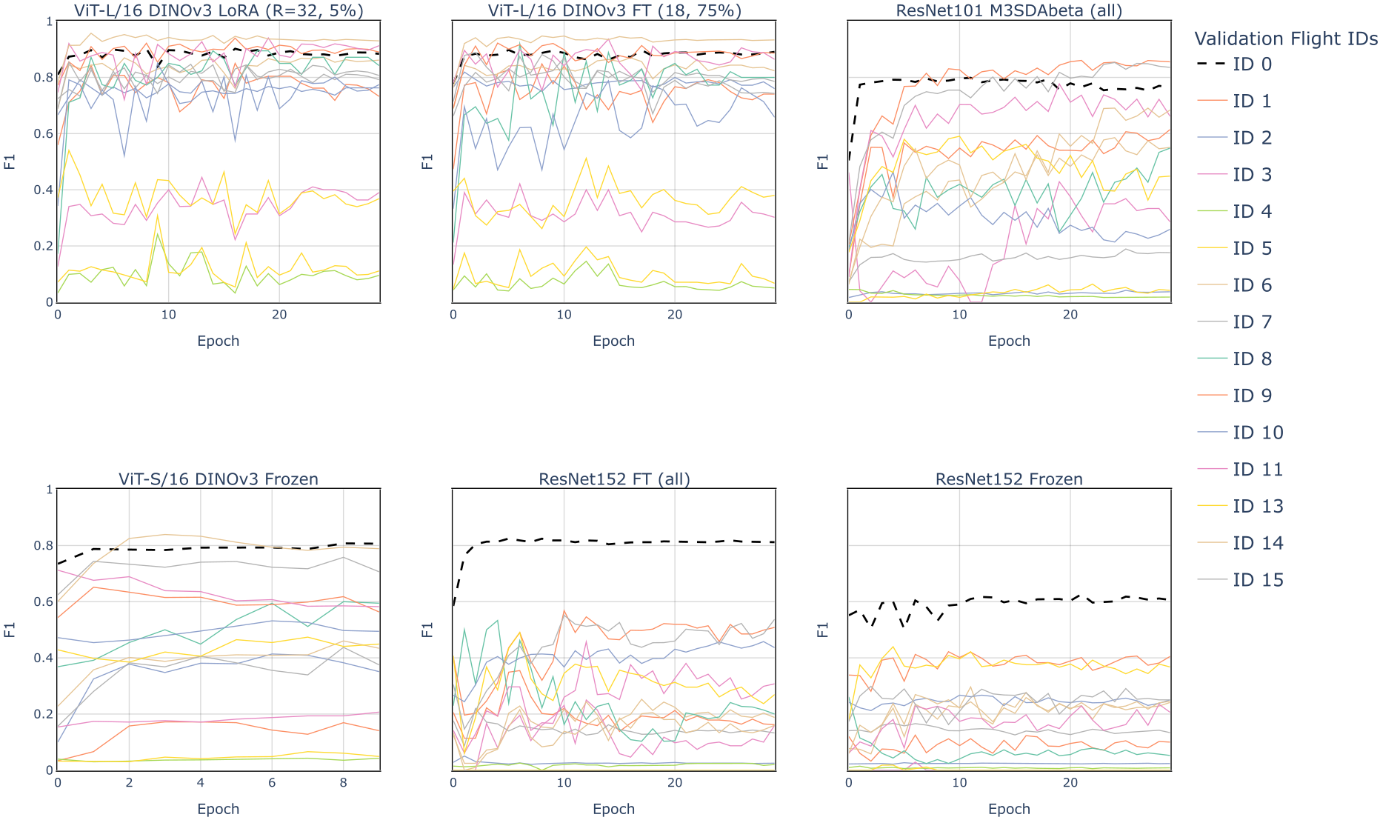}
\caption{Per-flight performance ($F_1$ scores versus epoch) for 6 different models.}\label{figSI1}
\end{figure}

$F_1$ scores versus training epoch are shown in Figure \ref{figSI1}. 

\section{Fine-tuning ResNet-based models}
\begin{figure}[thb]%% placement specifier
\centering%% For centre alignment of image.
\includegraphics[width=0.7\textwidth]{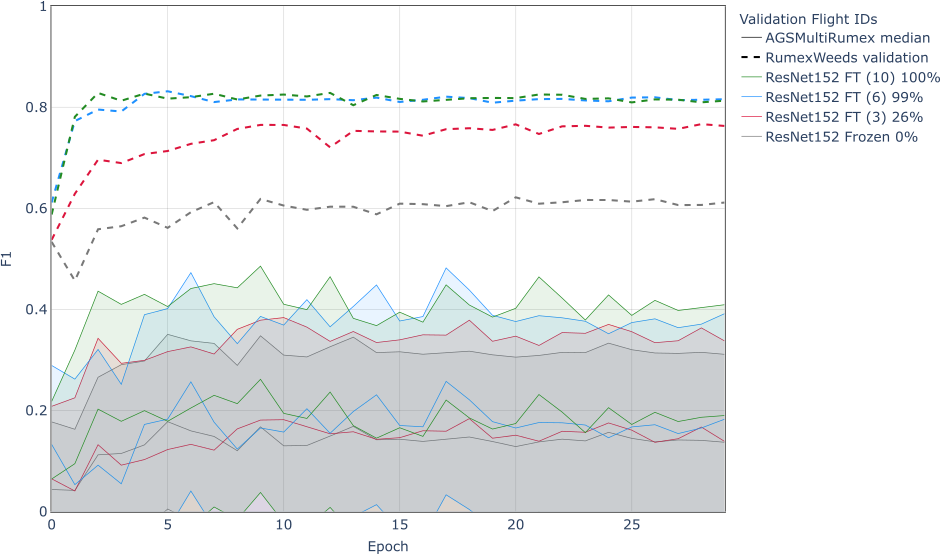}
\caption{$F_1$ score for ResNet-based models with different degree of vanilla fine-tuning (0, 3, 6 and 10 unfreezed blocks).}\label{figSI2}
\end{figure}

$F_1$ score versus training epoch for different ResNet-based models where fine-tuning involves 0, 3, 6 and 10 blocks, is shown in Figure \ref{figSI2}.

\section{False negative and true positve detections for flight Herrenpuent (ID 5)}
\begin{figure}[thb]%% placement specifier
\centering%% For centre alignment of image.
\includegraphics[width=\textwidth]{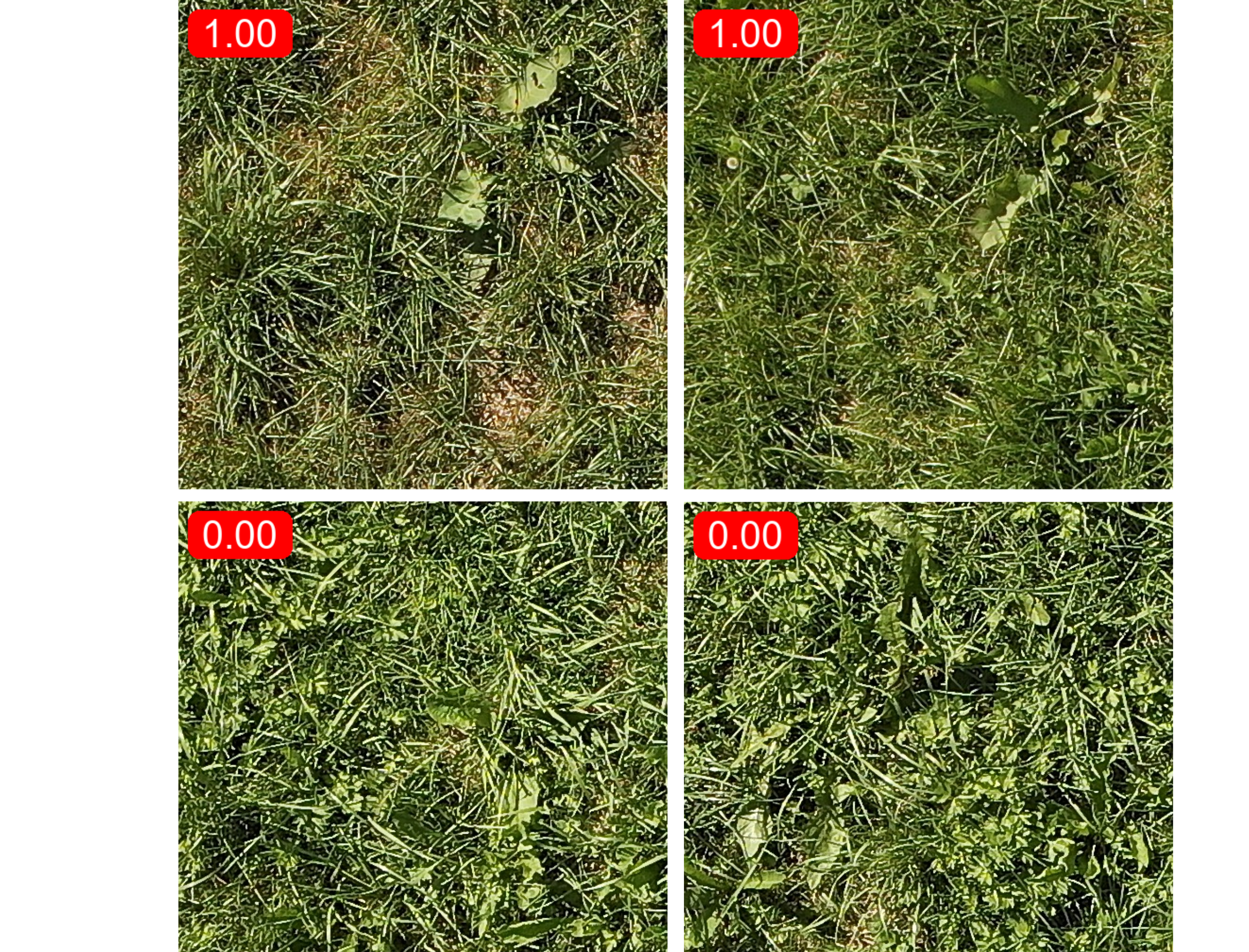}
\caption{Two examples of tiles correctly classified as Rumex (true positives, top) and tiles erroneously classified as background (false negatives, bottom) from flight Herrenpuent (ID 5) using the LoRA-finetuned ($R=32$) DINOv3 \textit{large}-based model. Number in red rectangle indicates uncalibrated model probability for class Rumex.}\label{figSI4}
\end{figure}

False negative detections remains challenging across all models studied for Herrenpuent (ID 5). Tile examples for these flights are shown in Figure \ref{figSI4}, together with the predictions of the best-performing model.

\section{False positive detections for different flights}
\begin{figure}[thb]%% placement specifier
\centering%% For centre alignment of image.
\includegraphics[width=\textwidth]{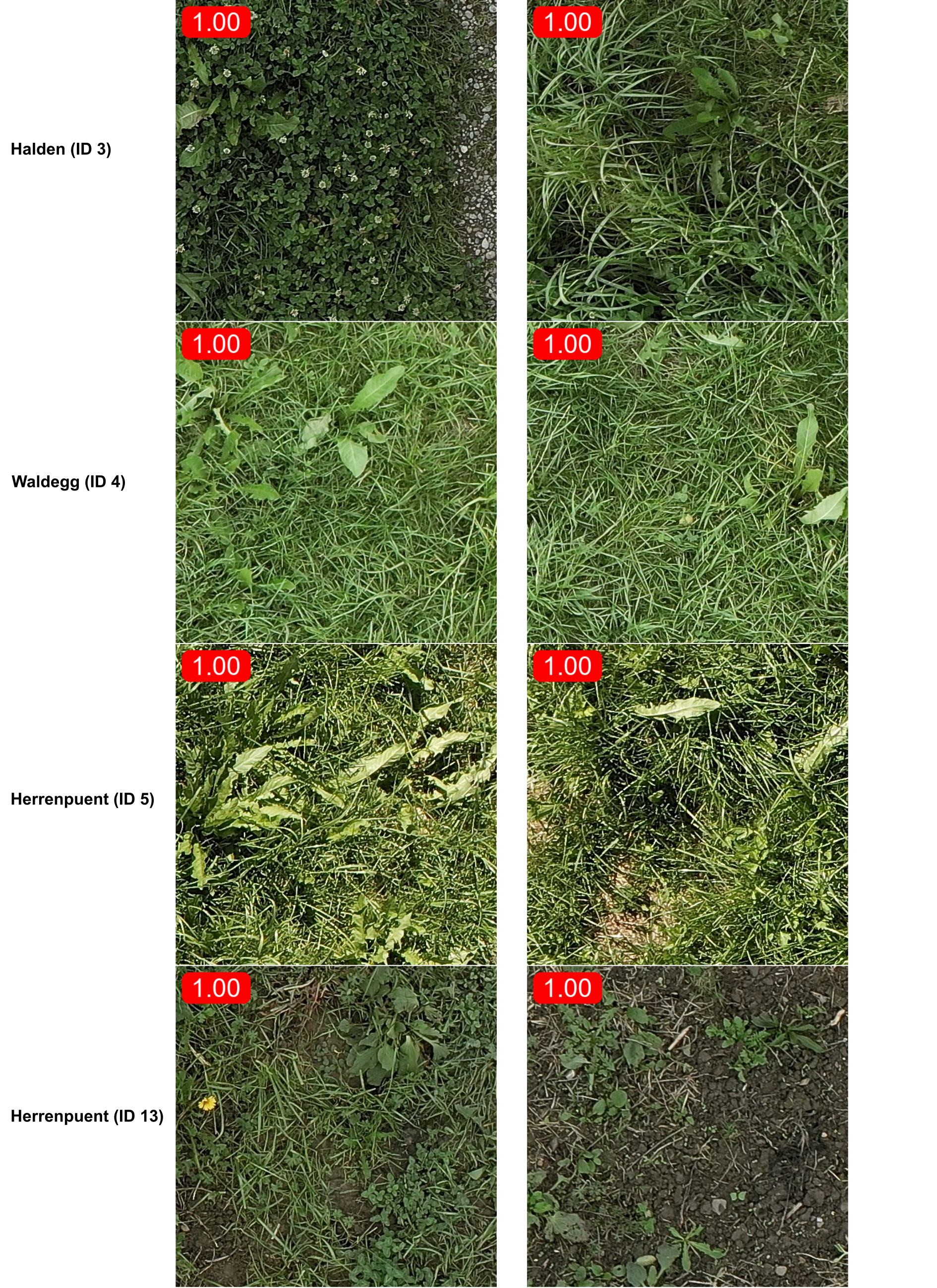}
\caption{Examples of background tiles erroneously classified as Rumex for flights Halden (ID 3), Waldegg (ID 4), Herrenpuent (ID 5) and Herrenpuent (ID 13). Model predictions were obtained using LoRA-finetuned ($R=32$) DINOv3 \textit{large}-based model. Number in red rectangle indicates uncalibrated model probability for class Rumex.}\label{figSI3}
\end{figure}

False positive detections remains challenging across all models studied, in particular for flights Halden (ID 3), Waldegg (ID 4), Herrenpuent (ID 5) and Herrenpuent (ID 13). Tile examples for these flights are shown in Figure \ref{figSI3}, together with the predictions of the best-performing model.

\end{document}